# Automatic Identification of Arabic expressions related to future events in Lebanon's economy

Moustafa Al-Hajj[1], Amani Sabra[2]

Lebanese University, Center for Language Sciences and Communication,
Celine Centre, Tayouneh, Beirut, Lebanon
[1]moustafa.alhajj@ul.edu.lb
[2]amani.sabra@hotmail.fr

**Abstract:** *In this paper, we propose a method to automatically identify future events in Lebanon's economy from Arabic texts. Challenges are threefold: first, we need to build a corpus of Arabic texts that covers Lebanon's economy; second, we need to study how future events are expressed linguistically in these texts; and third, we need to automatically identify the relevant textual segments accordingly. We will validate this method on a constructed corpus form the web and show that it has very promising results. To do so, we will be using SLCSAS, a system for semantic analysis, based on the Contextual Explorer method, and "AlKhalil Morpho Sys" system for morpho-syntactic analysis.*

**Keywords:** Arabic Natural Language Processing, Contextual Explorer, Lebanon's Economy, Corpus Construction.

## 1. Introduction

Due to the rapid growth of internet-based information, the need for filtering relevant information that is of specific interest to the user has become of paramount importance. In this paper, we consider identifying future events of Lebanon's economy from Arabic texts. Extraction of such information from trustworthy reports is of particular importance because they include information about Future Events in Lebanon's economy. Challenges are threefold: first, we need to build a corpus of Arabic texts that covers Lebanon's economy; second, we need to study how future events are expressed linguistically in these texts (in other words, what are the linguistic structures of future events); and third, we need to automatically identify the relevant textual segments conforming with these linguistic structures.

The study of future expressions considers that speech acts contain traces, on the surface, that can be used to automatically identify these expressions. Given this hypothesis, our objective is to show that specific structures can be used to identify future events in Arabic texts. Future expressions in Arabic have many common properties, making it possible to automatically identify them even in different speeches.

The forms which future expressions take in speech in various languages have been the study of many researches ([5], [1], [4]), and their automatic identification has been addressed in ([4], [1]). According to our knowledge, Al-Madyani in [6] is the only study in literature on Arabic expressions that predict the future in the Holy Quran and in the Arabic linguistic patrimony. Al-Madyani enumerated some linguistic methods and tools for making predictions in Arabic such as (1) methods using predictive particles like قد، لعل، عسى، ليت، هل …, (2) methods using predictive verbs and their derivatives like توقع، رجا … (3) methods using predictive structures like لم يفعل بعد (has not yet done), كان سيفعل (would have done) … (4) the use of "تَفَعَّل" form and (5) Contextual semantics methods. Despite the importance and the depth of the subject of this study, we noted the scarcity of its use in the Modern Standard Arabic (the modern version of classical Arabic), especially in press releases.

In this paper, we propose a method to automatically identify future expressions from a corpus we constructed of Arabic Web pages on Lebanon's economy. This method makes use of two systems, SLCSAS for semantic analysis which uses surface linguistic forms, and "AlKhalil Morpho Sys" for morphosyntactic analysis. In addition, we present a review of the SLCSAS system, as it is a new tool we developed based on the Contextual Exploration method [3], for automatic extraction and classification of sentences using a semantic map. We validate the proposed method on the constructed corpus and show that it achieves very promising results.

The outline of this paper is as follows. Section 2 presents a Web corpus construction method from Arabic web sources on Lebanon's economy. Section 3 describes linguistic structures of Arabic expressions related to future events. Section 4 describes the proposed method for automatic identification of Arabic expressions related to future events in Lebanon's economy. Finally, section 5 includes the evaluation and perspectives of this work.

## 2. Corpus construction method

We chose to work on constructing a corpus of texts on Lebanon's economy, because such a corpus is rich in information about future events in Lebanon's economy. A keyword search on the search engine Google.com involving the word list of Appendix 2, with the word "لبنان" (e.g. " الدين لبنان العامة "الموازنة ,لبنان "العام) led to the identification of relevant Web sources addressing the issue of Lebanon's economy. This list of keywords was selected from the "Economic and Social Reform Action Plan 2012-2015" report (http://www.pcm.gov.lb/Admin/DynamicFile.aspx?PHName=Document&PageID=2200&published=1), which proposes an economic matrix.

Figure 1 shows the components of the proposed corpus construction method from web sources. From the Google result pages in response to a query, only the first two pages are considered for automatically selecting URL addresses. Each page consists of 100 URLs at most. We arrived therefore at more than 1900 URLs of Arabic web pages, all of which were downloaded. Then, each HTML file was processed to identify the "main-article" content and title (between <title> tags), then the URL address is added to the result document that is to be compiled in the corpus. As such, each file of the corpus consists of the URL address, the title of the page and the main article content. The "main-article" content extraction is a hard problem to solve as well; however, we used some simple heuristics like extracting each sequence of letters (Arabic and Latin letters) and other characters (simple and double quotes symbols, coma, period, etc.) with a length greater than an empirical value (e.g. 130 characters). However, some articles may be duplicated; moreover, some unwanted content is obtained, therefore we continued improving content extraction heuristics to a point where we only get the main article. As such, out of the 1900 web pages, the content extraction of more than 1600 was relatively relevant and related to the corpus subject, giving us access to diverse Lebanese economic domains.

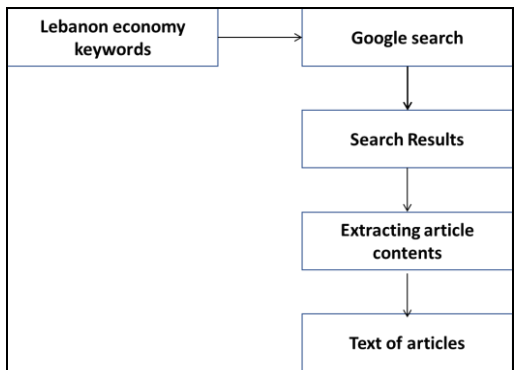

**Figure 1:** Corpus construction method components

## 3. Structures of Arabic future expressions

We will indicate with a few examples the linguistic forms associated with Arabic future expressions. Table 1 contains linguistic structures constructed around:

- The particle قد which means "it is possible, may be or might" when followed by a verb in the present tense [فعل مضارع], together indicate a future expression,
- The letter س which means "will" when prefixed to a verb,
- The particle لن which means will not or will never. It expresses the notion that something won't happen or occur in the future, and it is always followed by a verb in present tense,
- The particle سوف which means shall. It expresses the assertion that something will be realized in the future, and it is always followed by a verb in present tense,
- The predictive passive participles: ممكن (possible), متوقع (expected), مرجو (anticipated), مرجح (likely), مرتقب (desired), مستبعد (ruled out), محتمل (probable). All of them could be preceded by the linguistic structure: من الـ (e.g. من الممكن), which is made up of the Arabic definite article (ال) and the particle (من). In English, this could be expressed by using verb to be with the passive participle (e.g. It is expected to). These predictive participles can also be suffixed by the Aleph and Tanween (أ) diacritic (e.g. متوقعاً)
- The predictive verbs in past tense: توقع (expected), استبعد (ruled out), ارتقب (anticpated)
- The predictive verb in present tense: يتوقع (to expect), يستبعد (to rule out), يرجح (to consider likely), يرجو (to desire).

**Table 1:** Linguistic structures of Arabic future expressions

| Future expressions | Linguistic structures |
|---|---|
| Introduced with "قد" | ... قد [فعل مضارع] ... |
| Introduced with "س" prefixed to a verb | ... س[فعل مضارع] ... |
| Introduced with "لن" | ... لن [فعل مضارع] ... |
| Introduced with "سوف" | ... سوف [فعل مضارع] ... |
| Introduced with passive participles | ... (من ال)؟(ممكن| متوقع| مرجح| مرتقب| مرجو|مستبعد| محتمل)(أ)؟ ... |
| Introduced with verbs in the past | ... (توقع | ارتقب) ... |
| Introduced with verbs in the present | ... (يتوقع | يستبعد | يرجح | يرجو | ...) ... |

The examples provided here serve to illustrate different types of future expressions that have been identified:

- Expression with [فعل مضارع] قد

وقال "الحريري": ان لبنان يواجه مهمة صعبة متمثّلة في استضافة مليون ونصف المليون من اللاجئين السوريين، معتبراً ان "الخطر" الذي **قد يترتب** جراء عدم مساعدة لبنان في القيام بذلك لن ينعكس على اللبنانيين فحسب، بل "على العالم بأسره."

Hariri said: "Lebanon faces the difficult task of hosting one and a half million Syrian refugees", considering that the "danger" **that might result** from not helping Lebanon to do so will not only be reflected on the Lebanese, but also "on the entire world".

- Expression with [فعل مضارع]س:

فمثل هذا الاستثمار **سيوفر** عليه فاتورة سنوية تشكل اكثر من خمس فاتورة الواردات اللبنانية السنوية التي تزيد على 20 مليار دولار سنويا.

Such an investment **will provide** an annual bill representing more than one fifth of Lebanon's annual import bill of more than $ 20 billion a year.

- Expression with لن

واعتبر فرح ان استقالة الحريري "**لن** تؤدي بين ليلة وضحاها الى تغيير الوضعين المالي والاقتصادي، لان المصرف المركزي يمتلك احتياطيا نقديا ضخما يمكن استخدامه للدفاع عن الليرة اللبنانية."

Farah considered that Hariri's resignation "**will not** lead, overnight, to a change in the financial and economic situation, because the Central Bank has a huge cash reserve that can be used to defend the Lebanese pound."

- Expression with سوف

وعلى ضوء ذلك، يبدو ان الضغوط المالية التي يتعرض لها لبنان سوف تتزايد، وبما سيؤثر سلبا على استقرار الوضع النقدي والمصرفي في لبنان

Considering this, it appears that the financial pressures on Lebanon **shall increase**, and will negatively affect the stability of the monetary and banking situation in Lebanon

- Expression with predicting noun

فالازمة المستجدة **من المرجح** ان تثني وزارة المال عن اصدار سندات دولية كان مقررا اصدارها بالتعاون مع مصرف لبنان، بهدف ادارة الديون ودعم احتياطيات النقد الاجنبي.

The emerging crisis **is likely** to discourage the Ministry of Finance from issuing international bonds that were to be issued in cooperation with Banque du Liban, to manage debt and support foreign exchange reserves.

- Expression with a predicting agent noun

باسيل **مستبعدا** حدوث حرب كبيرة في سوريا: "لبنان ضد اي اعتداء على اي دولة عربية"

Bassil **ruled out** a major war in Syria: "Lebanon is against any aggression against any Arab country"

- Expression verb in past tense

**توقع** صندوق النقد الدولي أن يبلغ النمو الاقتصادي في لبنان 1.5 في المائة خلال عام 2017، متوقعاً أن يسجل لبنان وتيرة نمو ضعيفة خلال العام الحالي، نتيجة استمرار الأزمة السورية، التي لا تزال ترخي بظلالها وبشكل كبير على الاقتصاد والمجتمع المحليين.

The International Monetary Fund **expected** Lebanon's economic growth to reach 1.5 per cent in 2017, …

- Expression with verb in future tense

وبالاضافة الى ذلك، فانه **يتوقع** ان الخصخصة ستجلب المدخرات الجديدة وقد قلصت الوظائف الحكومية، وتخفض اسعار الفائدة، وحفز نمو القطاع الخاص الاستثمار الاجنبي.

In addition, privatization **is expected** to bring in new savings, cut government jobs, cut interest rates, and stimulate private sector growth in foreign investment.

## 4. Identification of Arabic future expressions

To automatically identify Arabic future expressions, we made use of two systems: SLCSAS (https://cslc.univ-ul.com/SLCSAS) and "AlKhalil Morpho Sys" (http://oujda-nlp-team.net/en/programms/alkhalil-morphology-2-en/). SLCSAS permits the automatic classification of sentences by analyzing surface linguistic forms. However, some structures need morpho-syntactic analysis, so we also made use of "AlKhalil Morpho Sys", a morphosyntactic analyzer system of Arabic, to identify verbs that refer to the future tense. Figure 2 shows the future expressions' identification process. First, all terms beginning with the letter س and those following the particle قد are identified, then they are analyzed by "AlKhalil Morpho Sys" to identify whether they refer to verbs in the future or not.

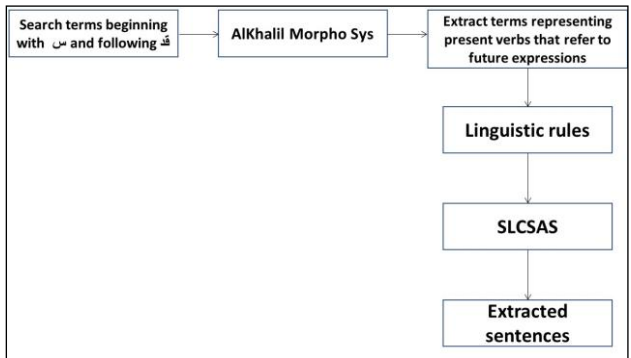

**Figure 2:** The future expressions' identification process

**Table 2:** SLCSAS linguistic rules for Arabic future expressions

| Comment | Linguistic Rule |
|---|---|
| "اسم_مفعول::" is a variable name representing a passive participle (see Appendix 1 for its values) In this rule and all following rules, "<-" "مستقبل" indicates the class name (Future in Arabic) | اسم_مفعول:: -> مستقبل |
| "فعل_مضارع_س::" is a variable name representing verbs in future tense beginning with the letter س (see Appendix 1 for some of its values ) | فعل_مضارع_س:: -> مستقبل |
| 5."(و\|ف)؟" means that the presence of one of the two letters (ف) or (و) is optional before قد 6."فعل_مضارع_قد::" is a variable name representing verbs in the future tense following the particle قد (See some of its values in Appendix 1) | (و\|ف)؟قد:: فعل_مضارع_قد -> مستقبل |
| "فعل_ماضي::" is a variable name representing verbs in the past tense (see values in Appendix 1) | فعل_ماضي:: -> مستقبل |
| "فعل_مضارع::" is a variable name representing verbs in the future tense (see values in Appendix 1) | فعل_مضارع:: -> مستقبل |
| "(و\|ف)؟" means that the presence of one of the two letters (ف) or (و) is optional before لن and سوف | (و\|ف)؟سوف:: -> مستقبل (و\|ف)؟لن:: -> مستقبل |

Then, the linguistic rules defined in the precedent section were rewritten in corresponding SLCSAS rule codes. A list of seven SLCSAS linguistic rules are defined as in Table 2.

For simplicity reasons, the future expressions are first identified, then the whole sentences thereof are extracted and thus referred to as future sentences. Following we present SLCSAS, the system we developed and used to identify future sentences.

### 4.1 SLCSAS

SLC Semantic Analysis System (SLCSAS) is a system developed using PERL at CSLC (Centre for Language Sciences and Communication) for the semantic classification of sentences following a given semantic map. It permits the extraction of sentences and their automatic classification by analyzing surface linguistic forms in their context. Moreover, it can be used to recognize entities in sentences. Only sentences satisfying a set of linguistic rules are extracted. The result consists of extracted sentences in HTML files with Linguistic markers highlighted in yellow for the retrieved positive markers. The red color is used in result to mark the research field of a negative marker. Figure 3 shows the architecture of the system.

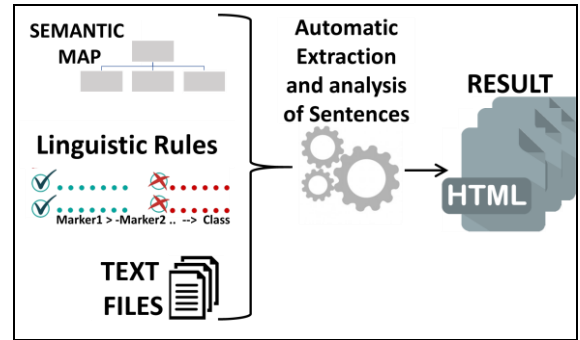

**Figure 3:** The architecture of SLCSAS

The classification of sentences is based on a set of linguistic rules. A Linguistic Rule (LR) consists of an ordered positive or negative Linguistic Form (LF), and a corresponding class. An LR takes the form as following:

*PM or -NM > PM or -NM > etc. -> SC*

where:
- *PM* = Positive Marker
- *NM* = Negative Marker
- *SC* = Semantic Category
- '-' symbol is used before each negative marker
- '>' symbol means 'followed by'
- After the '->' symbol comes the Semantic Category.

For a given text file in the corpus, each LR is tested on each sentence of the file to determine whether it applies to it. The testing process begins with the first LF (from the left for English or French LR and from the right for Arabic LR) and ends with the last one. The presence of each positive LF is mandatory in the search field. On the other hand, if a negative LF is present in the search field, the search process is canceled. As such, the system starts searching for of the first positive LF in the search field which is, at the beginning, the whole sentence. Then, it moves on to the second positive LF. The search field here is restricted to the part of the sentence that follows the first positive LF. Lengths (in number of words) of the search field of a positive or a negative LF are given in the system. The text segmentation into sentences is based on typographic forms like capital letters and dots; some heuristics are used depending on the input language. For example, for Arabic texts, a dot represents the end of sentence if it is followed by a space.

The analyzer was designed to accept, as an input, raw text files in UTF-8 encoding in different languages. The output of the system is a set of HTML files for articles showing extracted sentences grouped by semantic categories where positive markers are highlighted in yellow and the context of negative markers are highlighted in red. Negative markers appear when moving the mouse over parts highlighted in red. Furthermore, the system lets users choose which parts of the sentence they want to extract; the resulted excerpts, in this case, are underlined in the result.

When the title and the URL address of the web page are tagged in the input text, then the source and the title are identified and displayed in the result. Users can thus always read the full article offline or online (if it is still available on the original website) to further verify the correctness of the result.

The system is available for download, licensed under the GNU General Public License; the source is also included. The analyzer makes use of two files: "rules.txt" which contains all linguistic rules, and "semanticMap.txt" which contains the semantic map.

## 5. Evaluation and Perspectives

Results are available online in http://univ-ul.com/SLCSAS/ArFuturExpressions/results.html. The Figure 4 shows two screenshots of results from two files.

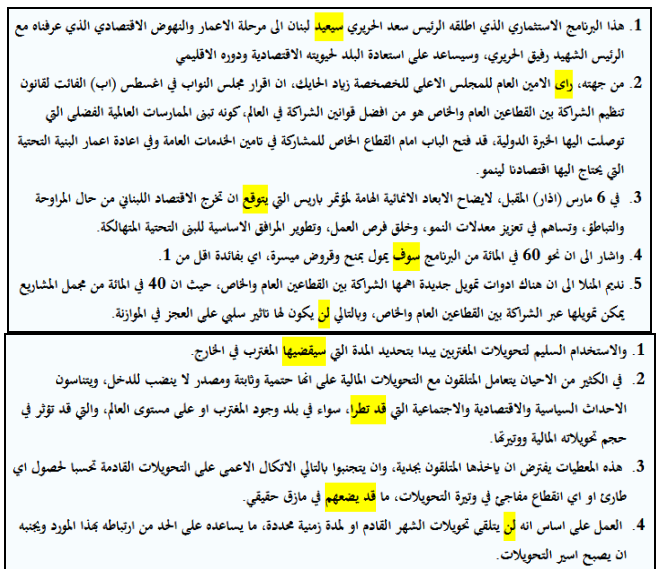

**Figure 4:** Snapshots of the results obtained from two web pages listing the automatically extracted future sentences. The examples show that all extracted sentences are true.

Based on our linguistic competence, we manually collected an exhaustive list of future sentences contained in a random collection of 200 files from the corpus. This list brings a total of 743 future sentences on Lebanon's economy, representing 16% of the whole 200 files sentences and their distribution per future sentence class is shown the Table 3.

**Table 3:** Distribution of future sentences in the Evaluation Corpus (200 files)

| Future sentence class | Number |
|---|---|
| Introduced with "قد" | 64 |
| Introduced with "س" prefixed to a verb | 450 |
| Introduced with "لن" | 93 |
| Introduced with "سوف" | 26 |
| Introduced with passive participles | 47 |
| Introduced with verbs in the past | 32 |
| Introduced with verbs in the present | 31 |

After the processing of all corpus files, 36,816 textual segments are identified as sentences, and 5,535 of them are identified as future sentences. According to the 200 randomly selected files, 4,634 textual segments are identified as sentences, and 762 of them are identified as future sentences, where 19 of them are incorrectly identified as future sentences, and all of the 743 true future sentences are identified. Table 4 summarizes the results obtained by our system. We used the Precision and the Recall measures for each future sentence class.

**Table 4:** Obtained Results for the various future sentence class

| Future sentence class | Precision | Recall |
|---|---|---|
| Introduced with "قد" | 94.11 | 100 |
| Introduced with "س" prefixed to a verb | 97.19 | 100 |
| Introduced with "لن" | 100 | 100 |
| Introduced with "سوف" | 92.85 | 100 |
| Introduced with passive participles | 100 | 100 |
| Introduced with verbs in the past | 100 | 100 |
| Introduced with verbs in the present | 100 | 100 |
| Overall | 97.50 | 100 |

Considering the simplicity of the proposed method, the overall results are very good, the system is fast and can detect all future expressions. However, some limitations of the proposed method are as follows:

- Some verbs in present tense following قد, in some cases, don't refer to future sentences, such as identified sentences with قد تجد (you may find), قد يشبه (might be like), قد يكون (it might be), قد يدل (may indicate), قد يعود (may be due).

  An example of a sentence extracted with 'due to' قد يعود:
  ان مؤشرات القراءة في الوطن العربي مرعبة وقد يعود ذلك الى انه يوجد حتى هذا التاريخ سبعون مليون امي
  The literacy indicators in the Arab world show terrifying figures, and this may be due to the fact that, to date, there are 70 million illiterates

- It is possible to have a word with the same form of (س + (فعل مضارع)), but referring to a proper noun, not a verb – such as سيدر, سيمون, سيشيل, etc. Similarly, the particle سوف might be preceded by (و) as in وسوف, which sometimes refers to a family name.

The present study focused on the identification of Arabic future expressions. In the future, we will consider working on the identification of the polarity of future expressions (positive, negative, neutral) based on word polarities. We will also consider working on the elaboration of the semantic map on Lebanon's Economy and the classification of future expressions.

## References

[1] Baeza-Yates, R. (2005). Searching the future. *In*: SIGIR Workshop on MF/IR.
[2] Boudchiche, M.; Mazroui, A.; Ould Abdallahi Ould Bebah, M.; Lakhouaja, A.; Boudlal, A.; 2017. *"AlKhalil Morpho Sys 2: A robust Arabic morpho-syntactic analyzer"*, Journal of King Saud University – Computer and Information Sciences. 29(2). pp. 141-146. DOI: 10.1016/j.jksuci.2016.05.002
[3] Desclés, J.-P., Jouis, C., Oh, H.-G., Reppert, D.: Exploration Contextuelle et sémantique: un système expert qui trouve les valeurs sémantiques des temps de l'indicatif dans un texte. In: Herin-Aime, R., Dieng, J.-P., Regourd, J.P. (eds.) Knowledge modeling andexpertise transfer, pp. 371–400. Amsterdam (1991)
[4] Lid, K. (2010). Future-referring expressions in English and Norwegian-a contrastive study based on the English-Norwegian parallel corpus. Available: http://www.duo.uio.no/
[5] Nakajima, Y, Ptaszynski, M, Honma, H, Masui, F, "Future Reference Sentence Extraction in Support of Future Event Prediction", International Journal of Computational Linguistics Research, Volume 9, Number 1, March 2018, pp. 27-41.
[6] سعد بن سيف المضياني، " أسلوب التوقع وطرائقه في العربية "، AlAthar Journal, Volume 27, Number 27, December 2016, pp. 199-212.

**Appendix 1:** Values of variable names

| Variable name | Values ("|" means or) |
|---|---|
| اسم_فاعل:: | (وإف)؟(من ال)؟( ممكن | متوقع | مرجح | مرتقب | مرجو | مستبعد ) |
| فعل_مضارع_س:: | سيفرض | وسنطلق | وسيجري | ستتمكن | سيجبره | سيفرح | ستصعب | سيوجهان | سيتقاضون | سيجبر | سنتر | ستجوب | سنمنح | ستتضمن ... |
| فعل_مضارع_قد:: | يواجهه | تستخدم | يعجز | يسفر | يستوعب | يطال | يطول | نلحظ | يشاهدها | تعيق | تتخلف | تعتقدين | تترتب | تستقر | تتراوح ... |
| فعل_ماضي:: | (وإف)؟(توقع | استبعد | ارتقب)(ت)؟ |
| فعل_مضارع:: | (وإف)؟(ي|ت|ن|ا)( توقع | ستبعد | رجح | رجو ) |

**Appendix 2:** List of keywords search

| Keyword translation | Keyword in Arabic | Keyword translation | Keyword in Arabic |
|---|---|---|---|
| Communication | الاتصالات | Economy | إقتصاد |
| health sector | قطاع الصحة | Government debt | الدين العام |
| Poverty reduction | الحد من الفقر | Money | المال |
| End of service | نهاية الخدمة | Tax | الضريبة |
| Education | التعليم | Public budget | الموازنة العامة |
| Women's Affairs | شؤون المرأة | Public property | الاملاك العامة |
| Youth | الشباب | Private sector | القطاع الخاص |
| Regional and local development | التنمية المناطقية والمحلية | Trade | التجارة |
| Municipalities | البلديات | Investment | الاستثمار |
| Corporation for Housing | الاسكان | Business Environment | بيئة الاعمال |
| Institutions and departments | المؤسسات والادارات | Public-Private Partnership | الشراكة بين القطاعين |
| E-government | الحكومة الالكترونية | Infrastructure | البنى التحتية |
| Civil Service Board | الخدمة المدنية | Energy | الطاقة |
| anti-Corruption | مكافحة الفساد | Oil and gas | النفط والغاز |
| Disaster Management | ادارة الكوارث | Transport | النقل |
| Tourism | السياحة | Water and Sanitation | المياه والصرف الصحي |
| Expatriate remittances | تحويلات المغتربين | Environment | البيئة |
| Passengers through the airport | | | المسافرين المطار |

## Author Profile

**Moustafa Al-Hajj** holder of a PhD degree in Computer Sciences from University of Tours, France. He received his B.S. and M.S. degrees in Computer Sciences from the Faculty of Sciences at the Lebanese University and University of La Rochelle - France in 2001 and 2003, respectively. Currently, Moustafa A-Hajj is an Associate Professor in Natural Language Processing in the Centre for Language Sciences and Communication at Lebanese University.

**Amani SABRA** holder of a PhD degree in "linguistics and data processing" from University Paris IV Sorbonne, France. She received her M.S. degree in LTLM/SIMIL "Lexicology Multilingual Terminology and Translation, Science d'Information Multilingue et Ingénierie Linguistique" from the Université Lumière Lyon II, France, and the B.S. degree in Science of Language and Communication from Faculty of Arts and Humanities at the Lebanese University. Currently, Amani SABRA is the director of the Centre for Language Sciences and Communication at Lebanese University and an associate professor of applied Linguistics and data processing at the Lebanese University.